\documentclass{article}

\usepackage{microtype}
\usepackage{subfigure}
\usepackage{booktabs} 
\usepackage[usenames,dvipsnames]{xcolor}
\usepackage{graphicx}
\usepackage{dcolumn}
\usepackage{bm}
\usepackage{siunitx}
\usepackage{textcomp}
\usepackage[utf8]{inputenc}
\usepackage{multirow}
\usepackage[version=3]{mhchem}
\usepackage{tcolorbox}
\usepackage{physics}
\usepackage{amsmath}
\usepackage{amsfonts}
\usepackage{amssymb}

\usepackage{hyperref}

\usepackage[accepted]{icml2020}

\icmltitlerunning{A Multiscale Graph Convolutional Network Using Hierarchical Clustering}

\begin{document}

\twocolumn[
\icmltitle{A Multiscale Graph Convolutional Network \\
           Using Hierarchical Clustering}




\begin{icmlauthorlist}
\icmlauthor{Alex Lipov}{to}
\icmlauthor{Pietro Lió}{to}
\end{icmlauthorlist}

\icmlaffiliation{to}{Department of Computer Science and Technology, University of Cambridge, UK}

\icmlcorrespondingauthor{Alex Lipov}{al822@cam.ac.uk}

\icmlkeywords{Machine Learning, ICML}

\vskip 0.3in
]



\printAffiliationsAndNotice{} 

\begin{abstract}
The information contained in hierarchical topology, intrinsic to many networks, is currently underutilised. A novel architecture is explored which exploits this information through a multiscale decomposition. A dendrogram is produced by a Girvan-Newman hierarchical clustering algorithm. It is segmented and fed through graph convolutional layers, allowing the architecture to learn multiple scale latent space representations of the network, from fine to coarse grained. The architecture is tested on a benchmark citation network, demonstrating competitive performance. Given the abundance of hierarchical networks, possible applications include quantum molecular property prediction, protein interface prediction and multiscale computational substrates for partial differential equations. 
\end{abstract}

\section{Introduction and Related Work}
\label{submission}
Most networks display some level of hierarchical topology \cite{barabasi2016network}; examples include actor networks, the semantic web and the internet at the autonomous system level \cite{PhysRevE.67.026112}. 

The defining feature of hierarchical network topology is how the clustering coefficient, $C_{i}$, of the $i$-th node scales with its degree, $k_{i}$; namely,
\begin{equation}
    C_{i}(k) \propto k_{i}^{-1}.
\end{equation}
Nodes with high degree and low clustering coefficient are those that connect different communities. Nodes with low degree and high clustering coefficient are those that connect within the community more than to different communities. Another distinct characteristic of hierarchical networks is that the clustering coefficient is independent of the number of nodes and that they show scale-free topology \cite{barabasi2016network}.

In this work, we present and explore a novel architecture operating on the graph domain which attempts to utilise the intrinsic hierarchical topological information that is embedded in a wide range of datasets. Graph neural networks, a booming field of deep learning, was built on the assertion that intrinsic graph structure in datasets was underutilised \cite{DBLP:journals/corr/abs-1812-08434}. However, the field of multiscale graph neural networks effectively does not exist, despite the fact that the evidence supporting the claim of hierarchical underutilisation is just as strong as for graph structure underutilisation \cite{PhysRevE.67.026112}.

Limited work has been done in the area of multiscale graph neural networks. Luzhnica, Day \& Liò \yrcite{DBLP:journals/corr/abs-1904-00374} introduced clique pooling; however cliques (complete subgraphs) are more restrictive than clusters (dense subgraphs) --- it is more likely that hierarchical networks consist of clusters rather than cliques (beyond triangular cliques). The work on multiscale graph convolutions by Haija \textit{et al.} \yrcite{DBLP:journals/corr/abs-1802-08888} is the closest piece of work to ours; however it does not use hierarchical clustering algorithms. 

In terms of applications; Lu \textit{et al.} \yrcite{lu2019molecular} used a hierarchical level-by-level treatment of quantum interactions in molecules, Zitnik \textit{et al.} \yrcite{zitnik2017predicting} used hierarchical multiplex graphs for multiple spatial scales of brain tissue and Kim \textit{et al.} \yrcite{san2019graph} developed a temporal multiscale graph convolutional neural network for analysing bike-sharing demands. Our multiscale decomposition was loosely inspired by the multiscale neural network using hierarchical block matrices, by Fan \textit{et al.} \yrcite{fan2019multiscale}, however it was designed to solve partial differential equations and does not operate on the graph domain.


\section{Architecture}

\begin{figure*}
    \centering
        \includegraphics[scale=0.142]{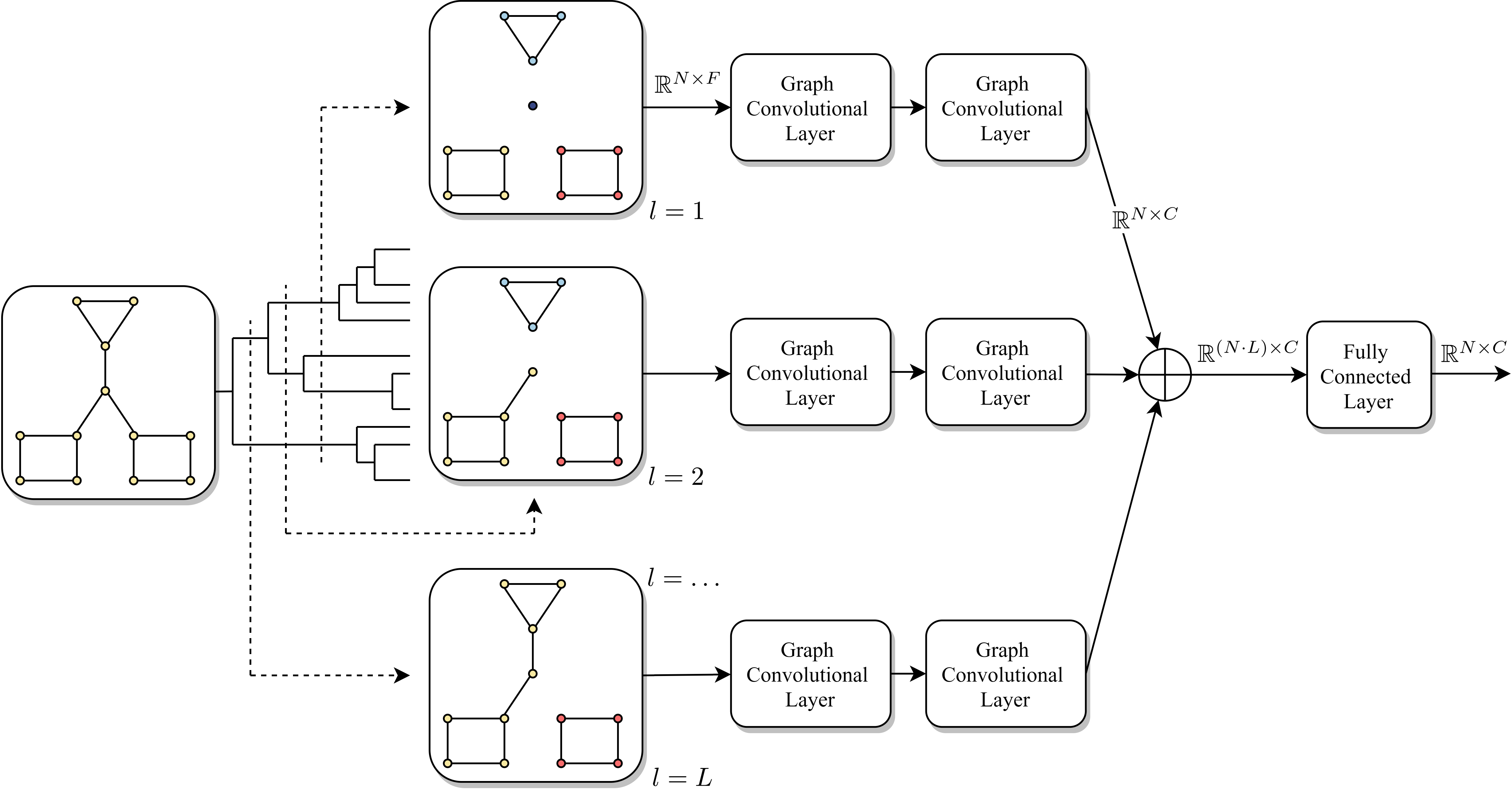}
    \caption{Our architecture which exploits hierarchical information through a multiscale decomposition. A dendrogram is produced by a Girvan-Newman hierarchical clustering algorithm, as shown on the left. It is segmented and fed through graph convolutional layers, shown in the middle, allowing the architecture to learn multiple scale latent space representations of the network, from fine to coarse grained. These representations are then combined, shown on the right, and fed through a fully connected layer which outputs the model's node classification predictions.}
\end{figure*}

The architecture is shown in Figure 1. We input an unweighted, undirected, graph, $\mathcal{G}=(\mathcal{V},\mathcal{E})$, where there are $N$ nodes, $\mathcal{V} \in \{\mathcal{V}_1, \dots, \mathcal{V}_N \}$, and $M$ edges, $\mathcal{E} \in \{\mathcal{E}_1, \dots, \mathcal{E}_M \}$. We construct $\hat{A}=A+I$, where $I \in \mathbb{R}^{N \times M}$ is the identity matrix and $A\in \mathbb{R}^{N \times N}$ is the adjacency matrix given by,
\begin{equation}
    A_{ij}=A_{ji}= \begin{cases}
    1   &\text{if node $i$ links node $j$} \\
    0   &\text{otherwise}
    \end{cases}.
\end{equation}
A multiscale decomposition, $f$, is taken through hierarchical clustering,
\begin{equation}
    f(\mathcal{G}) = \{\mathcal{G}_1, \mathcal{G}_2, \dots, \mathcal{G}_L \},
\end{equation}
and hence,
\begin{equation}
    f(\hat{A}) = \{\hat{A}_l \} \text{ where } l\in \{1, \dots, L\}. 
\end{equation}
We pass this and the input feature matrix, $X \in \mathbb{R}^{N \times F}$, with $F$ features per node, through the first convolutional layer, with $F$ input nodes, obtaining the set over $l$,
\begin{equation}
    \left\{ H_{l}^{(1)} \colon  H_{l}^{(1)} = \sigma\left( \hat{D}_{l}^{-\frac{1}{2}}\hat{A}_{l}\hat{D}_{l}^{-\frac{1}{2}} X W_{l}^{(0)}\right) \,  \right\},
\end{equation}
where $H_{l}^{(1)} \in \mathbb{R}^{N \times F^{'}}$, $F^{'}$ is the number of input nodes to the second layer, $W_{l}^{(q)}$ is the weight matrix for the $q$-th neural network layer on the $l$-th scale, $\sigma$ is a non-linear activation function and $\hat{D}_{l}$ is the $l$-th diagonal node degree matrix. The Kipf \& Welling \yrcite{DBLP:journals/corr/KipfW16} propagation rule was used in equation 5. We then pass through the second convolutional layer,
\begin{equation}
    \left\{ H_{l}^{(2)} \colon  H_{l}^{(2)} = \sigma\left( \hat{D}_{l}^{-\frac{1}{2}}\hat{A}_{l}\hat{D}_{l}^{-\frac{1}{2}} H_{l}^{(1)} W_{l}^{(1)}\right) \,  \right\},
\end{equation}
where $H_{l}^{(2)} \in \mathbb{R}^{N \times C}$, with $C$ being the number of classes. We then vertically concatenate the latent feature matrices from each scale, obtaining
\begin{equation}
    H^{(3)}=H_{1}^{(2)} \oplus H_{2}^{(2)} \oplus \dots \oplus H_{L}^{(2)}.
\end{equation}
Here, $H^{(3)} \in \mathbb{R}^{(N \cdot L) \times C}$. Finally this is fed through a fully connected (FC) layer, resulting in the node classification probability matrix as output, 
\begin{equation}
    Y = \sigma ( W H^{(3)} + b) ,
\end{equation}
where  $Y \in \mathbb{R}^{N \times C}$, the weight matrix $ W \in \mathbb{R}^{N \times (N \cdot L)}$ and $b$ is a bias matrix $ Y \in \mathbb{R}^{N \times C}$. In our implementation of this architecture, we chose $F^{'} = (F + C) / 2$. 

We used a graph convolutional network (GCN) layer depth of two here but this architecture can easily be generalised to any GCN layer depth. Ensembling, as we have done here, has significant literature demonstrating its power as an architectural choice \cite{xibin2020survey}.

\section{Experiments}

\begin{figure*}
    \centering
        \includegraphics[scale=0.55]{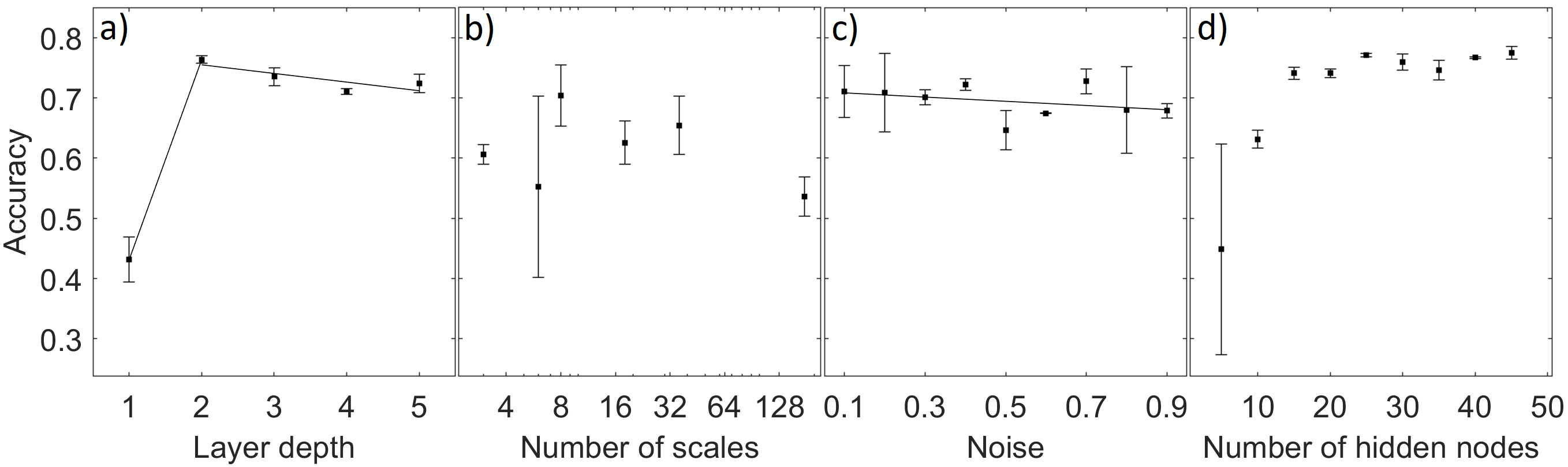}
    \caption{Node classification accuracy of the architecture when the GCN layer depth, number of decomposition scales and number of hidden nodes is varied. Accuracy upon corrupting the training set with random binary valued feature vectors is also shown, where the noise is altered. There is an optimal depth, shown by (a). Accuracy peaks at a threshold scale shown by (b), but this trend is somewhat weak. Robustness to noise is demonstrated by (c) and a threshold neural network width is required from observing (d).}
\end{figure*}

Tests were conducted on a truncated version of the Cora dataset (\url{https://relational.fit.cvut.cz/dataset/CORA}). Its properties are detailed in Table 1. Code can be found at [link removed for review].


\begin{table}[t]
\caption{Key properties of the truncated Cora dataset, used in the experiments.}
\vskip 0.15in
\resizebox{0.49\textwidth}{!}{%
\begin{tabular}{lr}
\toprule
Truncated Cora Dataset   &                                     \\ \midrule
Type                     & Machine learning citation network   \\
Task                     & Semi-supervised node classification \\
Number of nodes          & 2708                                \\
Number of labelled nodes & 140                                 \\
Number of edges          & 5429                                \\
Number of classes        & 7                                   \\
Feature vector length    & 1433                                \\ \bottomrule
\end{tabular}%
}
\label{Table 1}
\vskip -0.1in
\end{table}

\subsection{Hyperparameters}
Prior to performing the following experiments, the Cora network was put through a Girvan-Newman clustering algorithm. It produced 528 usable networks (excluding the last one which contained no edges, hence graph convolutions could not be performed properly). The non-linear activation function used was ReLU. Adam optimization was used with a learning rate of 0.01 and weight decay of $5 \times 10^{-4}$. The number of epochs used to train the convolutional layers was 300 for each graph. The number of epochs used to train the FC layer was 10. Given that this was a multi-class classification task with no major skew to the dataset, accuracy was deemed to be a sufficient performance metric. Two tests were performed for each data point, with error bars given by the standard deviation. The strategy for assigning the number of hidden nodes for a given GCN layer was to take the arithmetic mean of the number of nodes in the directly neighbouring layers, ensuring consistency. Each data point took approximately 10 minutes on a Nvidia GTX Titan X. Figure 2 summarises the results. Overall, the architecture achieved accuracies of 77\% using only 5\% of the truncated Cora dataset for training.

\subsubsection{Depth}
Three graphs were used. Graphs 0, 200 and 400 were used. That is, the coarse-grained representation, an intermediate scale and the finest scale  respectively. The number of FC hidden nodes was 30. The layer depth was varied from one layer to five layers. 

The results are shown in Figure 2(a). We note that the accuracy increases initially to its highest at layer depth 2, then drops steadily as more layers are added. This is consistent with a well-known phenomenon where below a problem specific threshold, the performance of the neural network is poor, but beyond that, more depth again leads to poor performance \cite{Loukas2020What}. It is speculated that a neural network that is too shallow is unable to learn more complex and abstract non-linear features, whereas a network that is too deep is liable to overfitting. Overly deep networks can also incur the vanishing gradient phenomenon \cite{ghosh2019exploring}.


\subsubsection{Scales}

Each graph from the multiscale decomposition can be interpreted as a different scale of the original network. GCN depth was kept at two. The number of FC hidden nodes was 7. We note this is less than for the layer depth experiment and we note the drop in accuracy. The number of scales used, $L$, was varied from 3 to 176. 

Figure 2(b) displays the results. There seems to be no strong trend. The highest accuracy was achieved with 8 scales. This was higher than the two lower number of scales, supporting the claim that multiscale decomposition does improve performance, but only up to a certain threshold number of scales. This is similar to the behaviour when altering the layer depth. 
\subsubsection{Hidden Nodes}
The GCN layer depth was kept at depth two. Three graphs were used: number 0, 200 and 400. The number of hidden nodes in the FC layer was altered from 5 to 45. The FC network always had 7 input nodes and 7 output nodes. 

The results are shown in Figure 2(d). The accuracy increases sharply when increasing from 5 to 20 hidden nodes, but then a plateau is reached, where increasing the number of hidden nodes does not significantly increase the performance. Individual more extreme tests, not plotted here, were performed at 100 and 1000 hidden nodes, both still returning approximately 76\% accuracy, confirming the plateau. Loukas \yrcite{Loukas2020What} suggests that the product of GNN width (number of hidden nodes) and depth has to exceed (a function of) the graph size to be performant. The neural network must be either deep or wide. This could explain the initial increase in accuracy as the number of hidden nodes is increased.
\subsection{Noise}

This experiment was conducted using standard hyperparameters: a GCN depth of two; the same three graphs from the depth and hidden node experiments; and 30 hidden nodes in the FC layer. The other experimental parameters detailed at the beginning of the hyperparameter section also apply here. 

The method for noise addition selected $140p$ nodes, out of 140 in the training set, where the noise parameter, $p \in [0.1, 0.9]$. It then replaced the feature vectors corresponding to those nodes with a random vector of binary values, maintaining the original dimensions. The values were sampled from a uniform probability distribution. 

The effect of varying the noise parameter on accuracy is shown in Figure 2(c). As should be expected, the accuracy decreases as the noise is increased. The model shows robustness.


\section{Discussion}

\subsection{Limitations}
There are three main limitations to the approach taken in the architecture described. First, there is added computational complexity due to the addition of a hierarchical clustering algorithm. The Girvan-Newman \yrcite{girvan2002community} algorithm scales as $\mathcal{O}(N^3)$ for a sparse network. The algorithm needs to be invoked anytime a new network is input into the model. It is therefore important to choose the optimal hierarchical clustering algorithm. Second, more RAM is needed to store the graphs produced by the multiscale decomposition. A memory efficient way of storing these could be devised, such as deserializing individual disk stored graphs, from the multiscale decomposition, each time one of the scales of GCN layers is trained, then removing them before training the next scale. Third, it is computationally more expensive to train many graph convolutional networks than training only one.  The performance benefit should outweigh the extra computation time.

\subsection{Future Work}
Performance could be compared across multiple datasets: a model artificially generated hierarchical network, a perfectly non-hierarchical network such as a random Erdős–Rényi \yrcite{erdHos1960evolution} graph and a stochastic block model network \cite{karrer2011stochastic} to understand the impact of adding communities. A neural network approach to hierarchical clustering could be implemented \cite{tian2014learning, yang2017graph}. Assessing performance while varying the distribution of scales, other than the linear spacing used here, may be informative. 

Architecture changes could include condensing nodes in each community to one node, rather than using graphs with different numbers of edges. The decomposition would produce graphs with different numbers of nodes; each cluster represented by a node with a feature vector obtained from averaging all the node features in that cluster. They would be pushed through graph convolutional layers, then the nodes would be unpacked by expanding each node into its original cluster nodes; this time the child nodes would all have the same feature vector from the parent node. This would be performed on all graphs and would allow the graphs to have the same number of nodes again which allows the usual flattening and FC layer.


\subsection{Applications}
First, the architecture would be well suited for quantum molecular property prediction using the QM9 or QM7b datasets \cite{blum, 1367-2630-15-9-095003, ramakrishnan2014quantum, ruddigkeit2012enumeration}. This is because of its hierarchical nature as demonstrated by the improved performance of a multilevel architecture by Lu \textit{et al}. \yrcite{lu2019molecular}. Quantum molecular property prediction can aid drug discovery. Second, the protein interface prediction network by Fout \textit{et al}. \yrcite{fout2017protein} could be augmented with our architecture, perhaps realising performance gains. This is a prime target for our architecture due to the hierarchical nature of protein interactions. 

Lastly, the graph element network by Alet \textit{et al}. \yrcite{alet2019graph} could be extended with our architecture. Graph element networks aim to be the neural network version of finite element analysis by using graph neural networks as computational substrates. Multiscale finite element analysis already exists where different scale meshes are used, from coarse to fine. Multiscale graph element networks do not exist, despite being a clear extension.



\bibliography{example_paper}
\bibliographystyle{icml2020}

\end{document}